\documentclass[10pt,twocolumn,letterpaper]{article}

\usepackage{cvpr}              

\usepackage[dvipsnames]{xcolor}

\definecolor{cvprblue}{rgb}{0.21,0.49,0.74}
\usepackage[pagebackref,breaklinks,colorlinks,citecolor=cvprblue]{hyperref}
\usepackage{multirow}
\usepackage{svg}

\newcommand\blfootnote[1]{
    \begingroup
    \renewcommand\thefootnote{}\footnote{#1}
    \addtocounter{footnote}{-1}
    \endgroup
}

\def\paperID{*****} 
\def\confName{CVPR}
\def\confYear{2024}

\title{What Appears Appealing May Not be Significant! - A Clinical Perspective of Diffusion Models}

\author{Vanshali Sharma\\
Indian Institute of Technology Guwahati\\
Assam, India\\
{\tt\small vanshalisharma@iitg.ac.in}
}

\begin{document}
\maketitle
\begin{abstract}
Various trending image generative techniques, such as diffusion models, have enabled visually appealing outcomes with just text-based descriptions. Unlike general images, where assessing the quality and alignment with text descriptions is trivial, establishing such a relation in a clinical setting proves challenging. This work investigates various strategies to evaluate the clinical significance of synthetic polyp images of different pathologies. We further explore if a relation could be established between qualitative results and their clinical relevance. 
\end{abstract}    
\section{Introduction}
In recent years, creating a pictorial view of what we imagine has become possible with the advent of diffusion models. This capability is coupled with text-based control, allowing us to type in a text prompt that describes our desired image output. These models can be fine-tuned to downstream tasks, expanding their viability to different applications. Generally, in most such applications, the visual outcomes during each fine-tuning iteration can be used to correlate them with improvement or overfitting. However, does this concept extend to medical images? Does solely relying on appealing visuals provide enough evidence for the clinical significance of generated medical images? These are some crucial questions that need attention for critical medical analysis tasks. 
\blfootnote{Accepted in WiCV (CVPR 2024) under poster category.}

In this work, we investigate the feasibility of synthetic medical data in both qualitative terms and clinical relevance to provide insights into the above questions. This study aims to generate colonoscopy images featuring polyps, which are precursors to colorectal cancer, the third most common malignancy. These polyps can be adenomatous (AD), which means malignancy potential, or non-adenomatous (Non-AD), which means benign. Additionally, we emphasized high-quality image generation. The synthetic images are obtained using a stable diffusion model~\cite{rombach2022high} and are further examined for pathological relevance by performing a binary classification (AD/Non-AD). The existing literature lacks such pathology-based studies and mainly focuses on polyp generation using binary masks \cite{machavcek2023mask,pishva2023repolyp}. Our contributions are summarized below:
\begin{itemize}
    \item We investigate the correlation between diffusion models' qualitative and quantitative results to determine their alignment or disparity in clinical settings.
    \item We explore the strategies that can be adopted to conclude the clinical relevance of synthetically generated images. 
\end{itemize}

\begin{figure}
    \centering
    \includegraphics[width=\columnwidth]{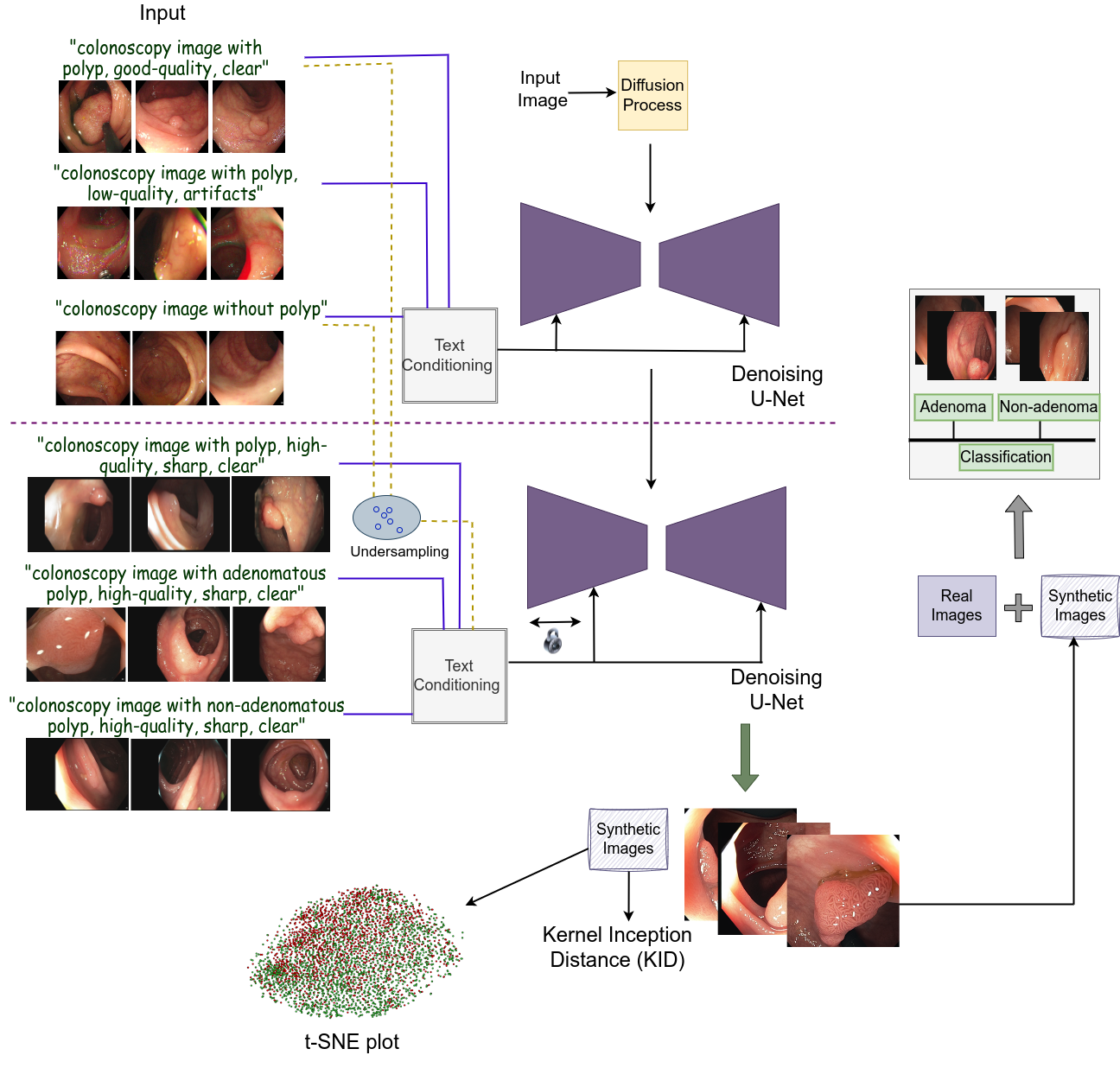}
    \caption{The image illustrates the diffusion model training process and assesses synthetic image quality in a clinical setting.}
    \label{fig:intro}
\end{figure}
\vspace{-10pt}

\section{Methodology}

\begin{table}[t]
\centering
\caption{Binary classification results using EfficientNetB0~\cite{tan2019efficientnet}. }
\resizebox{\linewidth}{!}{
\begin{tabular}{cccr@{\extracolsep{8pt}}rrrc}
\toprule                           & \multicolumn{3}{c}{Adenoma}                                                                                                               & \multicolumn{3}{c}{Non-Adenoma}                                                                                                          & \multicolumn{1}{c}{}                                     \\ \cline{2-4} \cline{5-7}
 \multicolumn{1}{l}{\multirow{-2}{*}{\begin{tabular}[c]{@{}l@{}}Training \\ images \end{tabular}}}     &  \multicolumn{1}{c}{Precision} & \multicolumn{1}{l}{Recall} & \multicolumn{1}{l}{F1-score} & \multicolumn{1}{c}{Precision} & \multicolumn{1}{l}{Recall} & \multicolumn{1}{l}{F1-score} &  \multicolumn{1}{l}{\multirow{-2}{*}{\begin{tabular}[c]{@{}l@{}}Balanced \\ Accuracy \end{tabular}}}   \\ \hline
  Real       &    \textbf{0.7901} &	\textbf{0.8312} & \textbf{0.8101} &	\textbf{0.5938} &	\textbf{0.5278} &	\textbf{0.5588} &	\textbf{0.6795}       			                                                       \\  Real+Synthetic & 0.7654 &	0.8052 &	0.7848 &	0.5313 &	0.4722 &	0.5000 & 0.6387 					                                                                                            				                                          \\
                    \bottomrule
\end{tabular}}
\label{tab:framecomp}
\end{table}
The overview of the proposed framework is shown in Fig. \ref{fig:intro}. Our approach adopts a stable diffusion based model with a denoising U-Net module. This module undergoes a two-step training process. Initially, the model learns from a large number of polyp and non-polyp colonoscopy images with additional quality-specific text prompts. This step allows it to learn polyp-focused patterns that could assist in further training it on a small-scale polyp classification dataset. Subsequently, we freeze some layers of the U-Net encoder and train it using pathology (AD/Non-AD) based high-quality images. To enhance the model's performance, we also included a few samples of the previous dataset and some high-quality images for which pathological annotations are unavailable. The generated images of AD and Non-AD classes are further evaluated using three strategies to assess their clinical relevance. These strategies include plotting t-SNE embeddings, using the Kernel Inception Distance (KID) metric, and augmenting the real images with generated images for a binary classification. Additionally, our approach incorporates iteration-wise analysis to ascertain whether a relation between qualitative outcomes and their clinical viability could be established.

\section{Experiments}
The experiments are conducted using three publicly available datasets, namely, SUN Database~\cite{misawa2021development} ($1,09,554$ non-polyp and $49,136$ polyp frames), CVC-ClinicHDSegment~\cite{vazquez2017benchmark,bernal2015wm,bernal2012towards} ($164$ images), and CVC-ClinicHDClassif~\cite{bernal2019gtcreator,sanchez2019computer} (train set: $536$ adenoma, $252$ non-adenoma, test set: $77$ adenoma, $36$ non-adenoma). We used the official validation set of CVC-ClinicHDClassif for testing due to the unavailability of the annotations in the official test set. The initial training of our diffusion model is conducted using the SUN Database, which is fine-tuned using the other two datasets. The pathological and clinical relevance evaluation is performed using CVC-ClinicHDClassif. The implementation uses the PyTorch framework, and execution is performed on NVIDIA A100 and NVIDIA Titan-Xp GPUs.

\begin{figure}
    \centering
    \includegraphics[width=\columnwidth]{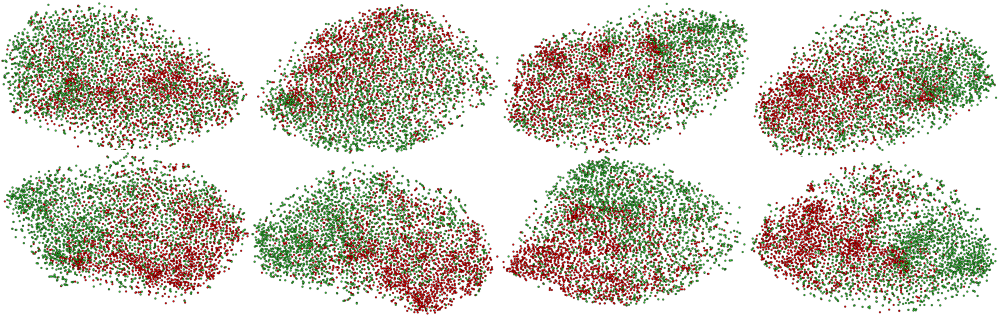}
    \caption{From left to right, and continuing the same sequence in the next row, the images correspond to t-SNE plots for iterations from 1k to 8k.}
    \label{fig:tsne}
\end{figure}

\begin{figure}
    \centering
    \includegraphics[height=80pt, width=0.7\columnwidth]{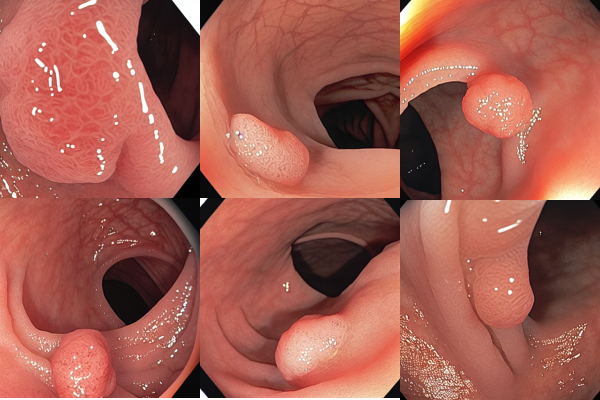}
    \caption{Row1: AD, Row2: Non-AD. From left to right, and continuing the same sequence in the next row, the synthetic images correspond to iterations 1k, 4k, and 8k.}
    \label{fig:sample}
\end{figure}

First, we conducted an iteration-wise analysis using t-SNE plots and the KID metric. Simultaneously, we tried to relate them with qualitative outcomes. It can be observed from Fig. \ref{fig:tsne}, Fig. \ref{fig:sample}, and Table \ref{tab:kid} that despite visually stunning outcomes in the initial iterations, the corresponding t-SNE plots and KID metrics signify clinical irrelevance of the synthetic images. This observation contradicts general images, where qualitative results often show correlations with quantitative outcomes. Moreover, it is difficult to visually conclude the pathology of such anomalies without clinical expertise. Furthermore, the best outcomes of t-SNE plots and KID metrics, i.e., iteration 8k, could not outperform real images when used for augmentation (see Table \ref{tab:framecomp}). The reason for underperformance can be inferred from Table \ref{tab:kid}, where iteration 8k is selected based on comparing synthetic AD/Non-AD with their real counterparts. However, it is important to consider the comparison between synthetic AD/Non-AD and real Non-AD/AD, respectively. This comparison signifies how distinct one class is from another, which is unfavorable in the 8k iteration.

\begin{table}[h!]
\centering
\caption{Iteration-wise quality assessment using KID.}
\label{tab:kid}
\resizebox{\linewidth}{!}{
\begin{tabular}{lllll}
\toprule
\textbf{Iteration} & \textbf{\begin{tabular}[c]{@{}l@{}}Synthetic AD \\ vs Real AD~$\downarrow$\end{tabular}} & \textbf{\begin{tabular}[c]{@{}l@{}}Synthetic AD\\  vs Real Non-AD~$\uparrow$\end{tabular}} & \textbf{\begin{tabular}[c]{@{}l@{}}Synthetic Non-AD \\ vs Real Non-AD~$\downarrow$\end{tabular}} & \textbf{\begin{tabular}[c]{@{}l@{}}Synthetic Non-AD \\ vs Real AD~$\uparrow$\end{tabular}} \\ \hline
1k                 & 0.073 (0.002)                                                               & 0.111 (0.003)                                                                   & 0.094 (0.003)                                                                       & 0.061 (0.001)                                                                   \\
2k                & 0.079 (0.002)                                                               & \textbf{0.118 (0.004)}                                                          & 0.100 (0.003)                                                                       & \textbf{0.067 (0.002)}                                                                   \\
3k                & 0.077 (0.002)                                                               & 0.114 (0.004)                                                                   & 0.087 (0.003)                                                                       & 0.061 (0.001)                                                                   \\
4k                & \textbf{0.069 (0.001)}                                                      & 0.106 (0.004)                                                                   & 0.085 (0.003)                                                                       & 0.057 (0.002)                                                                   \\
5k                & 0.074 (0.002)                                                               & 0.114 (0.004)                                                                   & 0.083 (0.003)                                                                       & 0.056 (0.002)                                                                   \\
6k                & \textbf{0.069 (0.002)}                                                      & 0.104 (0.003)                                                                   & 0.087 (0.003)                                                                       & 0.064 (0.002)                                                          \\
7k                & 0.076 (0.002)                                                               & 0.115 (0.003)                                                                   & 0.091 (0.003)                                                                       & 0.064 (0.002)                                                          \\
8k                & \textbf{0.069 (0.002)}                                                      & 0.105 (0.004)                                                                   & \textbf{0.081 (0.003)}                                                              & 0.061 (0.002)      \\
\bottomrule
\end{tabular}}
\end{table}

\section{Conclusion}
In this work, we proposed a diffusion-based approach to generate polyp images with different pathologies and investigated different strategies to assess their clinical relevance. We demonstrated that, unlike general images, it is difficult to establish a direct relation between qualitative outcomes and their clinical significance without some additional investigations. Our study provides pathways for future research using generative techniques with clinical images. 

{
    \small
    \bibliographystyle{ieeenat_fullname}
    \bibliography{main}
}


\end{document}